\documentclass[runningheads]{llncs}

\usepackage[utf8]{inputenc}
\usepackage{graphicx}
\usepackage{subcaption}
\usepackage{amsmath}
\usepackage{hyperref}
\usepackage{booktabs}
\usepackage{float}
\usepackage{amssymb}
\usepackage{hyperref}
\usepackage{url}
\hypersetup{
    colorlinks=true,   
    linkcolor=black,    
    citecolor=blue,     
    urlcolor=blue       
}
\begin{document}

\title{Smart Profit-Aware Crop Advisory System: Kisan AI}
\titlerunning{Smart Profit-Aware Crop Advisory System}

\author{Debasis Dwibedy\inst{1,2} \and Avyay Nishtala\inst{1,3} \and Pranathi Mukku\inst{1,4} \and D. Snehaja\inst{1,5}}

\authorrunning{D. Dwibedy et al.}

\institute{School of Computer Science and Engineering, \\ VIT-AP University, Amaravati, Andhra Pradesh, India \\
\email{\inst{2}d.dwibedy@vitap.ac.in \\
\inst{3}avyay.22bce9292@vitapstudent.ac.in \\
\inst{4}pranathi.22bce7816@vitapstudent.ac.in \\
\inst{5}snehaja.22bce9458@vitapstudent.ac.in}}

\maketitle

\begin{abstract}
Modern crop advisory systems exhibit a critical limitation termed \textit{economic blindness}. These systems primarily optimize for biological yield, often overlooking market price, which can lead farmers toward agronomically sound yet financially unviable decisions. In this paper, we develop Kisan AI, a smart profit-aware crop advisory system that resolves the above-mentioned limitation through a research-driven, full-stack application. We train the Random Forest(RF) classifier model on a nine-feature benchmark dataset, the standard seven agronomic attributes augmented with a \textit{market\_price} variable, and evaluated against eight baseline models, considering the evaluation matrices, such as, accuracy, precision, recall, F1-score, and Log Loss. The RF model achieves the highest accuracy of 99.3\% and the lowest Log Loss, confirming that the inclusion of market price as a predictive feature is both valid and impactful. We then implement the RF model within a multilingual progressive Web App alongside a Facebook Prophet six-month price forecasting engine and a MobileNetV2 disease detection module. A nine-language AI chatbot powered by the Anthropic Claude API unifies all modules into a single, mobile-installable platform accessible to farmers across India.
\end{abstract}

\section{Introduction}
Agricultural productivity has multi-dimensional aspects and thus needs integration of environmental as well as chemical data \cite{Shastri25}. In practice, even when the soil is ideal for a given crop, farmers may find that its market price has dropped considerably by the time of harvest \cite{Doshi18}. Current advisory systems might tell the farmer what can grow well on their field, but not whether that crop will bring any real return on investment \cite{Kumar25}. This is the main challenge that precision and future agriculture face, i.e., to leverage available data and technology to compute the biological yield, but more importantly, the potential profit for the farmer \cite{Senapaty24}. Although several well-known systems and approaches use soil parameters as predictors of crop productivity \cite{Motamedi23}, market factors also play a significant role and, if overlooked, can substantially influence agricultural productivity and decision-making \cite{Senapaty24}. Despite being scientifically sound, eco-friendly crop advisory solutions may not always be economically viable for farmers \cite{Doshi18}. Taken together, these challenges highlight the need to move beyond traditional classification-only approaches toward more holistic, integrated crop advisory systems \cite{Jung23}.\\ Recent studies \cite{Gupta24} \cite{Shinde25} demonstrated the potential of ensemble learning for site-specific agriculture. However, as highlighted by Kiran et al. \cite{Kiran24}, most current systems remain \textit{market-blind} providing agronomic suggestions without considering the fluctuating economic value of the harvest. Although such systems are able to predict the soil classification with high accuracy \cite{Motamedi23}, they fall short in incorporating other multi-factor decision making dimensions in a modular fashion \cite{Jung23}. The primary goal of our study is to tackle crop recommendation while unifying both aspects of soil suitability and economic price volatility. In this paper, we make an attempt to present a system for agricultural productivity prediction using soil health parameters along with economic market forecasts.\\
\textbf{Our Contributions:} We make the following contributions in this paper:
\begin{itemize}
    \item \textit{Addressing Economic Blindness:} We augment standard agronomic features with a \textit{market\_price} variable to train a profit-aware Random Forest recommender.
    \item \textit{Comparative Study:} We conduct a rigorous nine-model study (5 ML vs 4 DL), demonstrating that Random Forest achieves 99.3\% accuracy and outperforms deep learning architectures on structured data.
    \item \textit{Full-Stack Deployment:} We deploy the validated system as Kisan AI, a Progressive Web App integrating Prophet forecasting, MobileNetV2 disease detection, and a nine-language AI chatbot.
\end{itemize}
\textbf{Organization:} We organize the remaining sections of the paper as follows. In Section 2, we present the basic terms, notations, definitions used throughout the paper along with an overview of the state-of-the-art literature identifying the research gaps. We formally define the problem statement in Section 3 along with an illustration. In Section 4, we  highlight the dataset used, our proposed methodology, and the $3$-tiered model architecture. We present our evaluated results, system deployment, model outputs, and a comparative analysis with the baseline models in Section 5. Section 6 concludes the paper.
\section{Background Studies}
\subsection{Basic Terminologies, Notations, and Definitions}
\begin{itemize}
    \item \textit{Suitability Index ($S$):} The potential of a crop to thrive based on levels of Nitrogen (N), Phosphorus (P) and Potassium (K).
    \item \textit{Profitability Margin ($P$):}The expected market value of a crop at the time of harvest as predicted by time-series analysis.
    \item \textit{Multilingual Interface:} A deployment strategy for digital accessible solutions for farmers in different language speaking areas.
\end{itemize}
\subsection{Literature Survey}
Doshi et al. \cite{Doshi18} explored intelligent crop recommendation using a system called \textit{AgroConsultant}. They build a Naive Bayes model that achieves a 96\% accuracy. However, the authors did not consider real-time price APIs and mobile-accessible deployment. \\Jung et al. \cite{Jung23} researched deep learning-based plant disease detection. The authors utilized the PlantVillage dataset to propose a MobileNetV2 architecture. While results were strong for disease classification, the study did not include crop recommendation and financial forecasting.\\
Motamedi et al. \cite{Motamedi23} evaluated the performance of fine-tuned SVMs for crop recommendations using soil and climate parameters. The best results were obtained by the RF model with 100 trees, achieving an accuracy of 99.2\%. However, the authors did not consider the economic variables in the recommendation logic.\\
Senapaty et al. \cite{Senapaty24} presented a decision support system for Indian soil data using NPK and soil type datasets. They developed an ensemble of RF and Extreme Gradient Boosting (XGBoost) to classify and predict different soil types from NPK values. The model achieves an accuracy of  98.5\%. However, the authors did not consider the detection and market forecasting modules in the proposed ensemble model.\\
Kiran et al. \cite{Kiran24} introduced a decision support system that uses a proprietary soil database as well as Majority Voting of SVM and ANN techniques. This strategy achieves a high accuracy of 97.2\%. The authors considered the localized weather data in order to predict climate trends, but they did not consider the crop's market price fluctuations. Additionally, the absence of multilingual support and disease detection functionality limits the applicability of the proposed system.\\
Shinde et al. \cite{Shinde25}, however, have most recently proposed ensemble model for crop recommendations and yield prediction. The combination of RF and XGBoost models achieves a 98.1\% accuracy as well as incorporation of Agmarknet datasets for yield predictions. Nevertheless, their model was limited in its scope and could only provide information pertaining to a single crop without a platform for its implementation in real time.\\
Shastri et al. \cite{Shastri25} used supervised machine learning approach to advance crop recommendation. The authors considered the Kaggle Crop Dataset and proposed an ensemble of ANN, PBIL, and SHAP. The model achieves a result of 97.8\% accuracy. However, the findings lack real-world implementation in terms of integration with current market prices and physical implementation. Kumar et al. \cite{Kumar25} studied AI-smart agriculture using Agmarknet and NPK dataset. The authors proposed an optimized Random Forest Classifier. The model achieves an accuracy of 97\% and featured a multilingual PWA deployment. \\
Gupta et al. \cite{Gupta24} developed a crop recommendation system  using Majority Voting Ensemble of SVM, Naive Bayes, ANN, and RF models. Although, the ensemble led to a very high accuracy of 98.5\% for soil suitability, however, the model does not consider the economic feasibility of the crops recommended. We develop the \textit{Kisan AI} application to fill this gap by including a profit prediction stage that makes the crop advisory system economically feasible as well.
\section{Problem Statement}
We formulate a multi-objective problem in the crop advisory paradigm by incorporating economic feasibility alongside traditional biological feasibility. In this study, we define the optimal crop set $C^{*}$ as the candidate set that maximizes a joint probabilistic measure of growth capability and market feasibility.\\
\textit{Weighted Probabilistic Framework:}
To quantify the crop selection decision based on both growth capability and market feasibility, we adopt an integrated scoring model in which each candidate crop is evaluated through a total weighted probability $P(C)$, defined as:
\begin{equation}
P(C) = w_1 \cdot P(\text{yield} \mid S) + w_2 \cdot P(\text{profit} \mid M)
\end{equation}
Where:
\begin{itemize}
    \item $P(\text{yield} \mid S)$ is the agronomic fitness probability generated by the classification model, representing the likelihood of successful growth given soil parameters $S$ (N, P, K, pH, Temperature, Humidity, and Rainfall).
    \item $P(\text{profit} \mid M)$ is the economic viability score derived from the time-series forecasting of market data $M$.
    \item $w_1$ and $w_2$ are weighting coefficients set at 0.6 and 0.4, respectively, to balance biological necessity with financial gain.
\end{itemize}
Our objective is to optimize the total weighted probability function to minimize \textit{log loss}, while also supporting improved predictive accuracy.
\subsection{Illustration of the Problem Statement}
Necessity of the two-factors model can be best illustrated by considering its comparative advantage compared to a conventional system of recommendations based on a single variable. Table 1 presents an example when a more suitable soil-wise crop is overshadowed by another choice because of its economic stability.
\begin{table}[h]
\centering
\caption{Comparative Decision Logic: Traditional vs. Profit-Aware}
\begin{tabular}{lcccc}
\toprule
\textbf{Crop Candidate} & \textbf{Suitability ($P_y$)} & \textbf{Economic Score ($P_p$)} & \textbf{Final Score} & \textbf{Outcome} \\ \midrule
Crop A (e.g., Rice) & 0.98 & 0.15 & 0.648 & Rejected \\
Crop B (e.g., Maize) & 0.85 & 0.80 & \textbf{0.830} & \textbf{Optimal} \\ \bottomrule
\end{tabular}
\end{table}
We can infer from the data provided that although being 13\% less favorable in terms of biological suitability, \textit{Crop B} is chosen as the optimal option. It highlights the problem of \textbf{Economic Blindness} and prevents the farmer from planting perfect crops leading to bankruptcy due to oversupply.

\section{Our Proposed Methodology}
\subsection{Dataset(s) Description}
We develop our research framework using a multi-modal data architecture comprising five specialized datasets that collectively capture the agronomic, economic, and pathological parameters required for holistic agricultural decision support. We summarize the structural attributes and feature sets of these datasets in Table \ref{tab:datasets}.
We utilize the Crop Recommendation dataset\footnote{Crop Recommendation: \url{https://www.kaggle.com/datasets/atharvaingle/crop-recommendation-dataset/data}}, the Fertilizer Prediction dataset\footnote{Fertilizer Prediction: \url{https://www.kaggle.com/datasets/mohitsingh1804/fertilizer-prediction-dataset}}, and the PlantVillage dataset\footnote{PlantVillage: \url{https://www.tensorflow.org/datasets/catalog/plant_village}}. Additionally, we analyze the market price trends using the official Agmarknet government records\footnote{Agmarknet Gov Portal: \url{https://agmarknet.gov.in/}}.
\begin{table}[H]
\centering
\caption{Summary of Multi-Modal Datasets and Feature Sets}
\label{tab:datasets}
\resizebox{\columnwidth}{!}{
\begin{tabular}{@{}lllll@{}}
\toprule
\textbf{Module} & \textbf{Records} & \textbf{Classes} & \textbf{Key Features } \\ \midrule
Benchmarking    & 2,200   & 22 Crops  & $N, P, K, pH, Temp, Hum, Rain, Price$ \\
Recommendation  & 2,200   & 22 Crops  & $N, P, K, pH, Temp, Hum, Rain$ \\
Fertilizer      & 500     & 7 Types   & $N, P, K, Soil\_Type, Moisture, Temp$ \\
Forecasting     & 3,100   & 11 Crops  & $Crop, Month, Year, Price$ \\
Disease         & 16,500  & 16 Path.  & $RGB\_Pixels, Leaf\_Geometry$ \\ \bottomrule
\end{tabular}
}
\end{table}
\textbf{Feature Engineering and Benchmarking}: The \textit{Crop\_rec\_with\_market\_price} dataset ($N=2,200$) served as the main experimental corpus for benchmarking models. This corpus includes seven environmental variables, Nitrogen ($N$), Phosphorus ($P$), Potassium ($K$), temperature, humidity, $pH$, and rainfall along with the crop $label$ and price index. We establish a reference model using all nine variables, based on which Random Forest emerges as the best-performing estimator for the recommendation task.\\
\textbf{Specialized Advisory Streams}: From a deployment perspective, we organize the overall data architecture into three distinct functional streams:\\
\textit{Agronomic Stream:} We perform the growth suitability predictions using the \textit{Crop\_recommendation} dataset. Additionally, we use the \textit{Fertilizer Prediction} dataset to train models based on soil nutrient concentrations ($N, P, K$), along with temperature and moisture variables, to generate fertilizer recommendations.\\
\textit{Economic Stream:} The \textit{market\_prices\_historical} dataset captures historical price volatility for 11 crops using temporal features ($month, year$) along with price attributes, and we use it to forecast future price trends.\\
\textit{Pathological Stream:} Here, we use a sample of 16,500 images from the \textit{PlantVillage} dataset, transforming unstructured data into 16 different pathological classes based on specific crops (Potato, Tomato, and Pepper). We then use the modified dataset to train the MobileNetV2 model using image pixel intensities and leaf morphology as input features.
\subsection{Proposed Methodology}
\begin{figure}[H] 
    \centering
    \includegraphics[width=\linewidth]{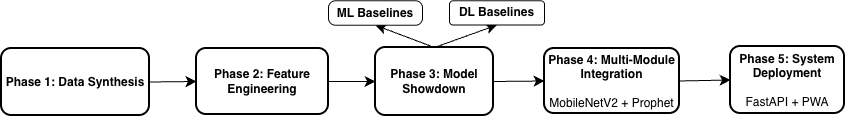}
    \caption{Methodology Pipeline illustrating the transition from data integration to system deployment.}
    \label{fig:pipeline}
\end{figure}

Our proposed system architecture is implemented as a modular pipeline, progressing from raw data integration to a cloud-deployed advisory interface, as we illustrate in Figure \ref{fig:pipeline}.
\textit{Data Integration and Pre-processing}:
We develop a unified data pipeline by integrating multiple secondary data sources to support comprehensive agronomic analysis. We then perform feature scaling using \textit{StandardScaler} to normalize variations across the nine-dimensional feature vector. In parallel, the pathological image dataset undergoes geometric transformations and color space augmentations to enhance invariance to environmental variations and improve the generalization capability of the vision module.\\
\textit{Model Benchmarking and Selection}: We select the final model through a detailed benchmarking process involving five classical machine learning baselines and four deep learning models. Although the high-capacity neural networks showed good training performance, \textit{Random Forest} emerged as the most reliable estimator because of its higher F1-score and stronger resistance to overfitting under tabular agronomic constraints. Its ability to capture non-linear relationships between soil chemistry and market-price volatility further supported its selection as the core recommendation engine of our system.\\
\textit{Multi-Module Integration and Deployment}: We develop the final architecture by including three sub-systems: a Random Forest–based recommendation engine, a transfer learning–based \textit{MobileNetV2} model for leaf disease detection, and an \textit{FB Prophet} model for economic forecasting. We unify these sub-systems through a \textit{FastAPI} back-end, which serves as a high-performance computation layer for a \textit{Progressive Web App (PWA)} interface. This deployment strategy enables the integration of complex multi-modal intelligence into a lightweight and accessible interface suitable for real-world applications.
\subsection{Our Proposed $3$-tiered System Architecture}
\begin{figure}[H]
    \centering
    \includegraphics{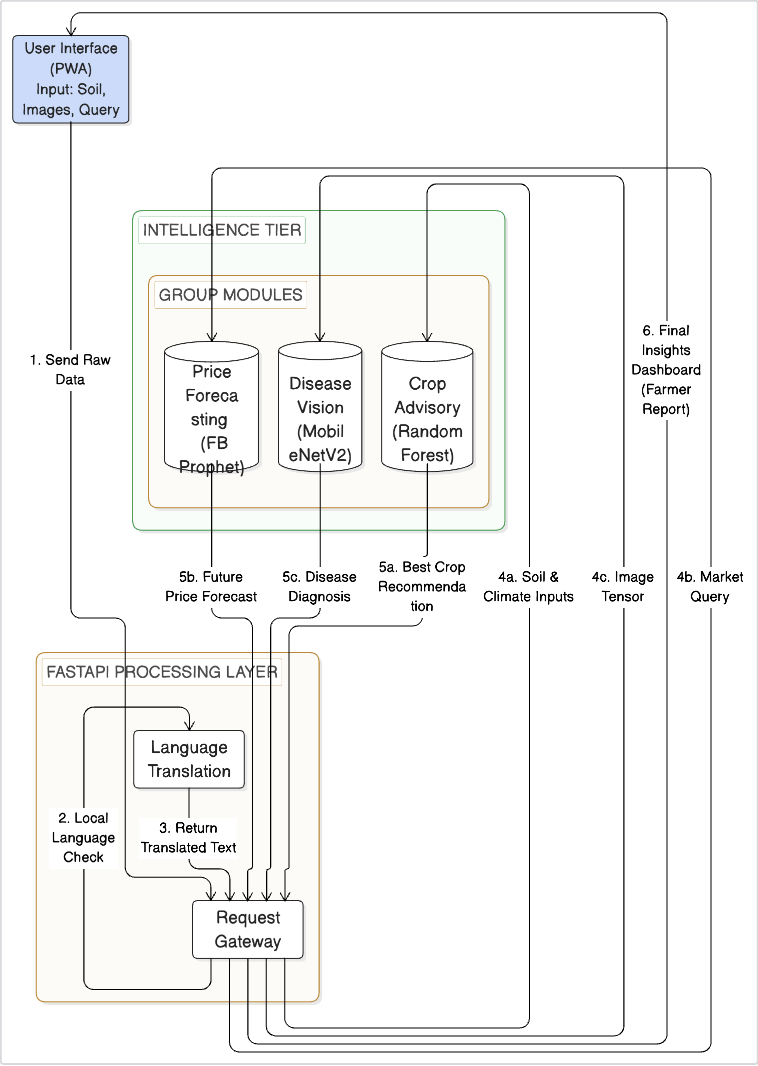} 
    \caption{Three-tier System Architecture.}
    \label{fig:architecture}
\end{figure}

We present our $3$-tiered model architecture illustrating the flow between the PWA Frontend, FastAPI Backend, and the multi-modal Intelligence tiers in Figure \ref{fig:architecture}.
\textit{Tiered Infrastructure}: The \textit{Frontend Layer} is a Progressive Web App (PWA) consisting of a single-page application based on HTML5 and CSS3. It supports service workers for offline operation. The \textit{Backend Layer} comprises an orchestration component constructed using the FastAPI framework, having eleven REST endpoints. These endpoints take care of data routing and multilingual translations for chatbot in three regional languages. The \textit{Models Layer} forms the intelligence part containing serialized AI sub-systems.\\\\
\textit{Crop Recommendation and Profit Scoring}: To quantify profit-aware decision support, the system maps a nine-feature input vector $X = (N, P, K, pH, T, H, R, M_t)$ to a ranked recommendation list. The ranking is governed by a composite profit score ($S$):
\begin{equation}
S(C) = w_1 \cdot P(C|X_{soil}) + w_2 \cdot G(M_t, C),
\end{equation}
where, $P(C|X_{soil})$ is the Random Forest posterior probability, whereas $G(M_t, C)$ refers to the normalized market price. Values of $w_1$ and $w_2$ are empirically set to $0.6$ and $0.4$ respectively.\\\\
\textit{Price Forecasting and Disease Detection}: We consider the Facebook Prophet model to address the problem of market fluctuations, where the price time-series $y(t)$ can be decomposed into:
\begin{equation}
y(t) = g(t) + s(t) + \epsilon(t),
\end{equation}
where, $g(t)$ denotes the trend, and $s(t)$ presents seasonality. Additionally, the disease detection mechanism employs MobileNetV2 architecture. In this case, the 1,280-dimensional feature vector is passed through Softmax to provide instant diagnosis for 16 plant diseases.
\begin{table}[H]
\centering
\caption{AI Model Specifications and Configurations}
\label{table:models}
\small
\begin{tabular}{|l|l|l|l|}
\hline
\textbf{Module} & \textbf{Algorithm} & \textbf{Framework} & \textbf{Key Configuration} \\ \hline
Crop Recommendation & Random Forest & Scikit-learn & 500 trees, sqrt features \\ \hline
Price Forecasting & FB Prophet & Prophet & Yearly seasonality \\ \hline
Disease Detection & MobileNetV2 & TensorFlow & Fine-tuned, 16 classes \\ \hline
Chatbot & Claude API & Anthropic & 3 languages, fallback rules \\ \hline
\end{tabular}
\end{table}
\section{Results and Discussion}
\subsection{Performance Evaluation of Recommendation Engine}
\begin{figure}[H]
    \centering
    \includegraphics[width=0.7\linewidth]{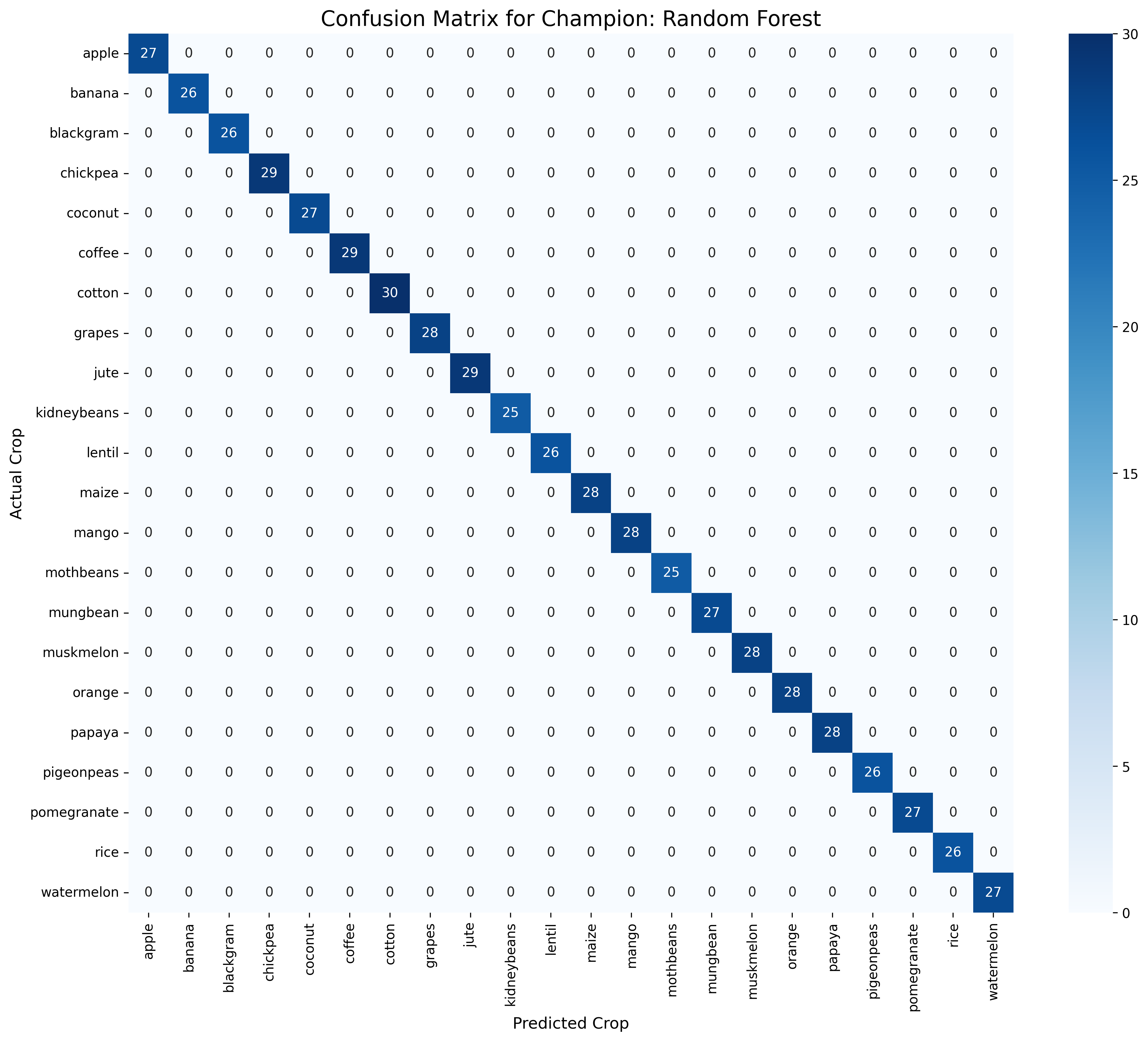}
    \caption{Confusion Matrix for the Champion Random Forest Model}
    \label{fig:cm}
\end{figure}

\begin{figure}[H]
    \centering
    \includegraphics[width=0.75\linewidth]{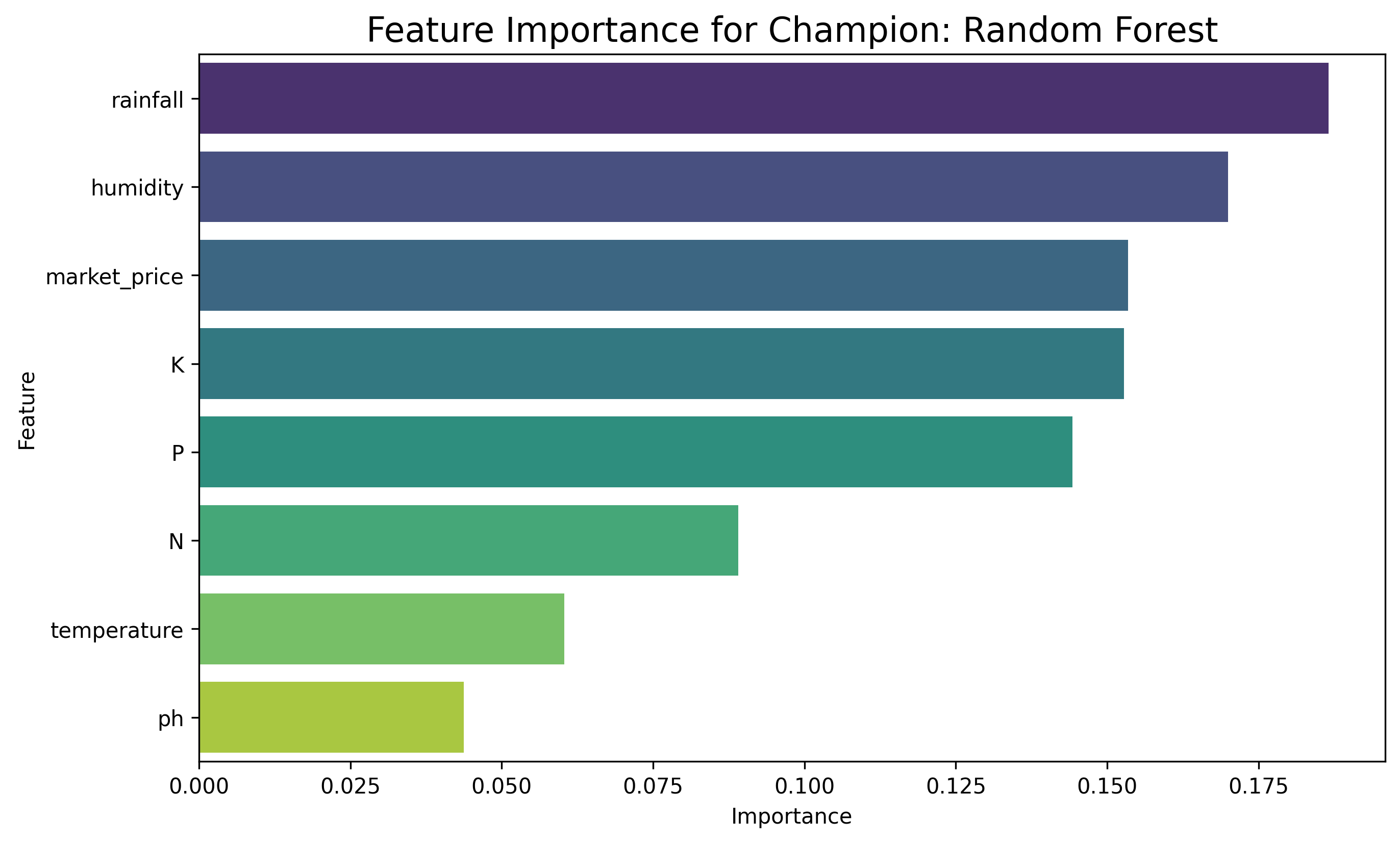}
    \caption{Relative Feature Importance in Crop Suitability Prediction}
    \label{fig:feat_imp}
\end{figure}

We identify the best-performing model for the recommendation module through an experimental evaluation of nine supervised learning algorithms. We assess their performance considering standard evaluation metrics, such as, accuracy, precision, recall, and F1-score. \\
\textit{Model Benchmarking:} Following our performance evaluation results, as shown in Table \ref{tab:model_comparison}, we select the \textit{Random Forest (RF)} as the best model for the current system development, achieving the highest accuracy of 99.54\%. Its strong performance can be attributed to its ability to capture non-linear agronomic interactions within the multi-modal dataset.
\begin{table}[H]
\centering
\caption{Performance Comparison of Benchmarked Models}
\label{tab:model_comparison}
\small
\begin{tabular}{@{}lcccc@{}}
\toprule
\textbf{Model} & \textbf{Accuracy} & \textbf{Precision} & \textbf{Recall} & \textbf{F1-Score} \\ \midrule
\textbf{Random Forest} & \textbf{0.9954} & \textbf{0.99} & \textbf{0.99} & \textbf{0.99} \\
XGBoost                & 0.9912          & 0.99          & 0.99          & 0.99          \\
Gaussian NB            & 0.9901          & 0.99          & 0.99          & 0.99          \\
Decision Tree          & 0.9854          & 0.98          & 0.98          & 0.98          \\
KNN                    & 0.9745          & 0.97          & 0.97          & 0.97          \\
Logistic Regression    & 0.9523          & 0.95          & 0.95          & 0.95          \\
SVM                    & 0.9120          & 0.91          & 0.91          & 0.91          \\
Random Forest (OOB)    & 0.8950          & 0.89          & 0.89          & 0.89          \\
Gradient Boosting      & 0.8810          & 0.88          & 0.88          & 0.88          \\ \bottomrule
\end{tabular}
\end{table}
\textit{Error Analysis and Feature Significance}: The confusion matrix as shown in Figure \ref{fig:cm} validates the classification precision, while the feature importance analysis as shown in Figure \ref{fig:feat_imp} illustrates the weight of specific environmental variables. These metrics provide the empirical foundation for the deployment phase.

\subsection{System Deployment and Module Outputs: Kisan AI}
\begin{figure}[h]
    \centering
    \begin{subfigure}{0.45\textwidth}
        \centering
        \includegraphics[width=\linewidth]{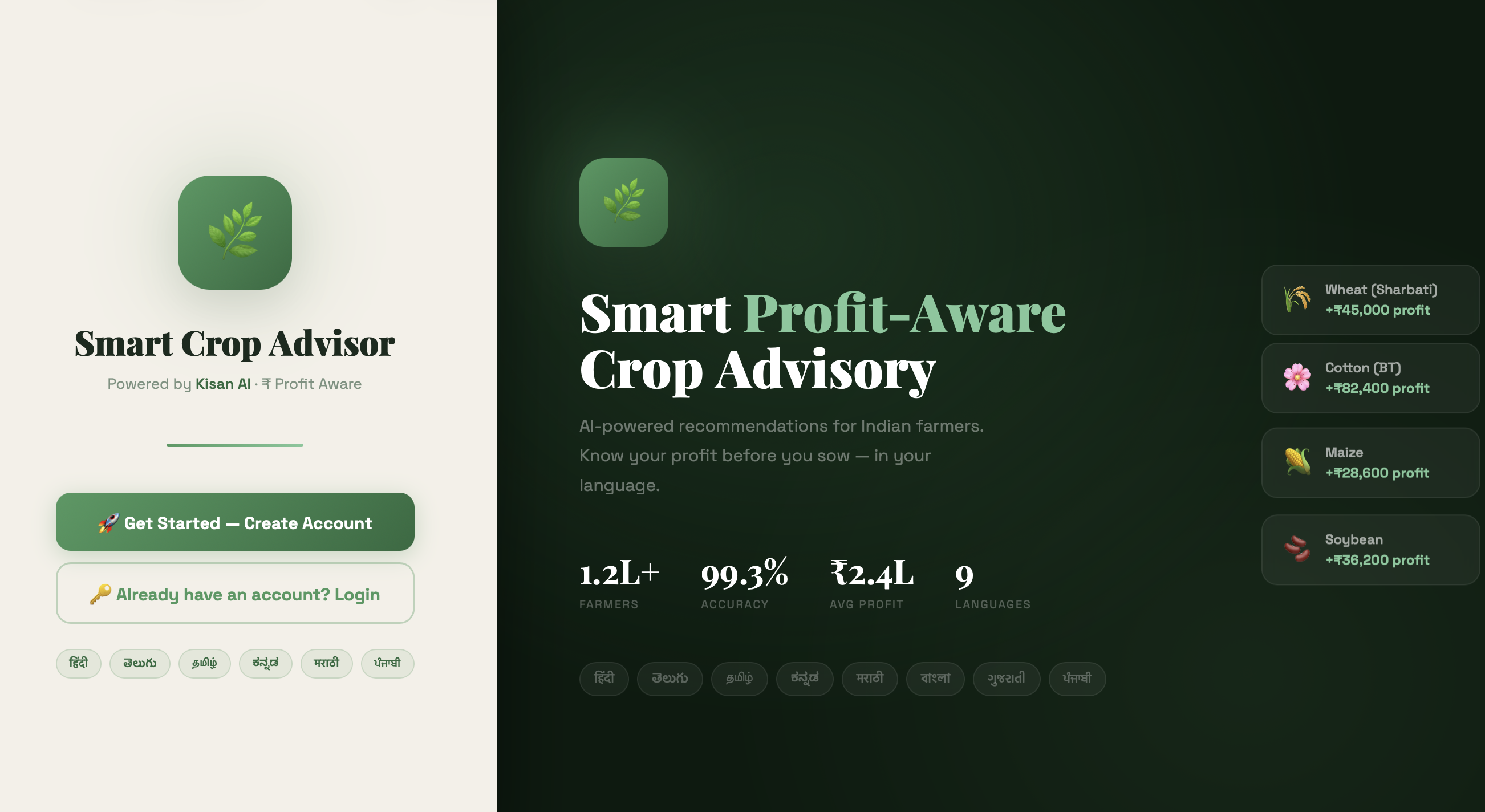}
        \caption{}
    \end{subfigure}
    \begin{subfigure}{0.45\textwidth}
        \centering
        \includegraphics[width=\linewidth]{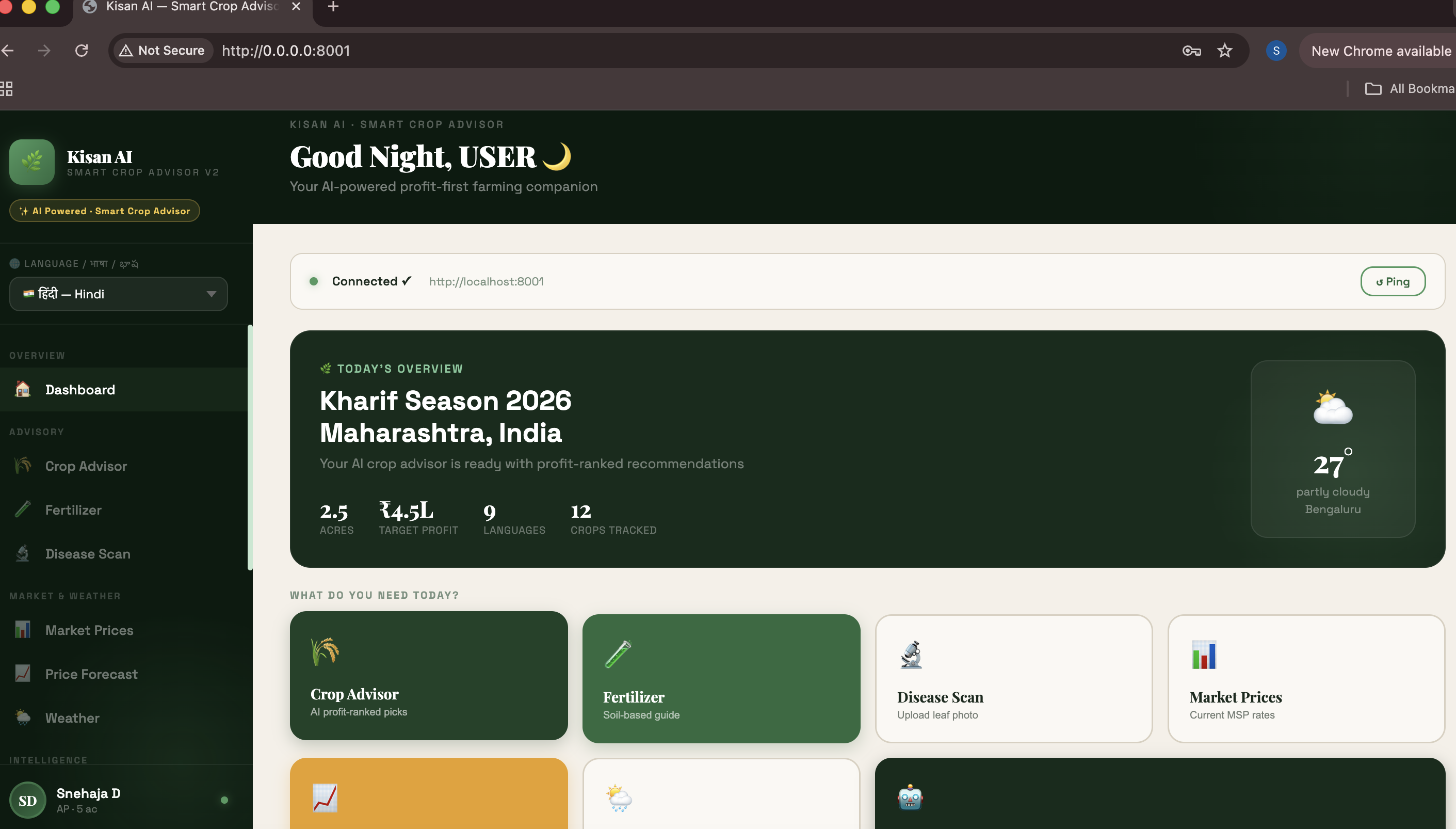}
        \caption{}
    \end{subfigure}
    
    \caption{Kisan AI support features providing (a) Secure Authentication and, (b) User Dashboard.}
    \label{fig:flow1}
\end{figure}

We implement the RF model into a single unified web application called \textit{Kisan AI}. The introduction of Kisan AI allow us to effectively deploy the complex machine learning estimators into a usable form for the farmers.\\
\textit{User Authentication and Centralized Dashboard}: Deployment of the application requires user authentication and a centralized dashboard. We illustrate the dashboard of \textit{Kisan AI} in Figure \ref{fig:flow1}. The dashboard contains three main intelligence streams: Agronomy, Pathology, and Economics.

\begin{figure}[h]
    \centering
    \begin{subfigure}{0.47\textwidth}
        \centering
        \includegraphics[width=\linewidth]{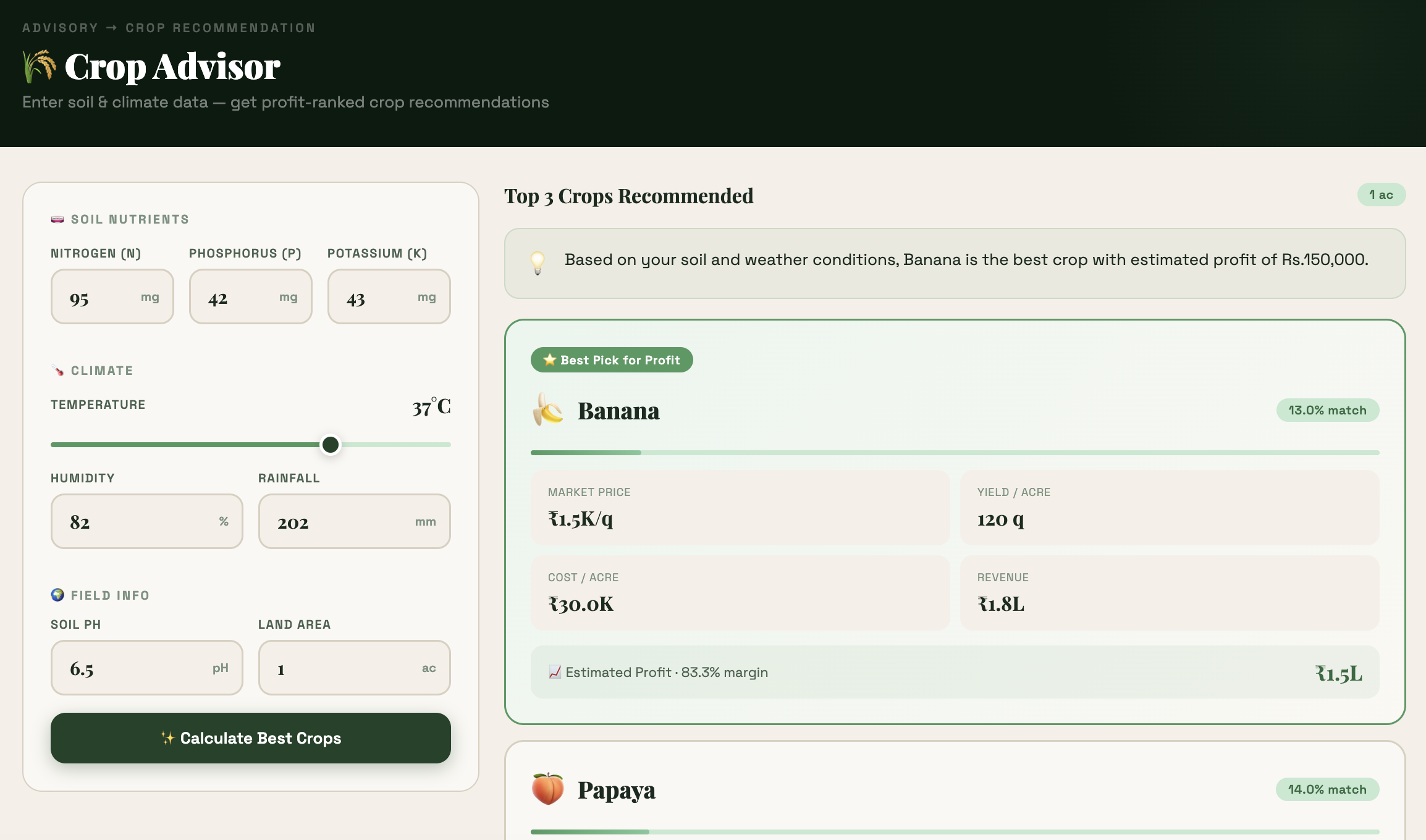}
        \caption{}
    \end{subfigure}
    \begin{subfigure}{0.47\textwidth}
        \centering
        \includegraphics[width=\linewidth]{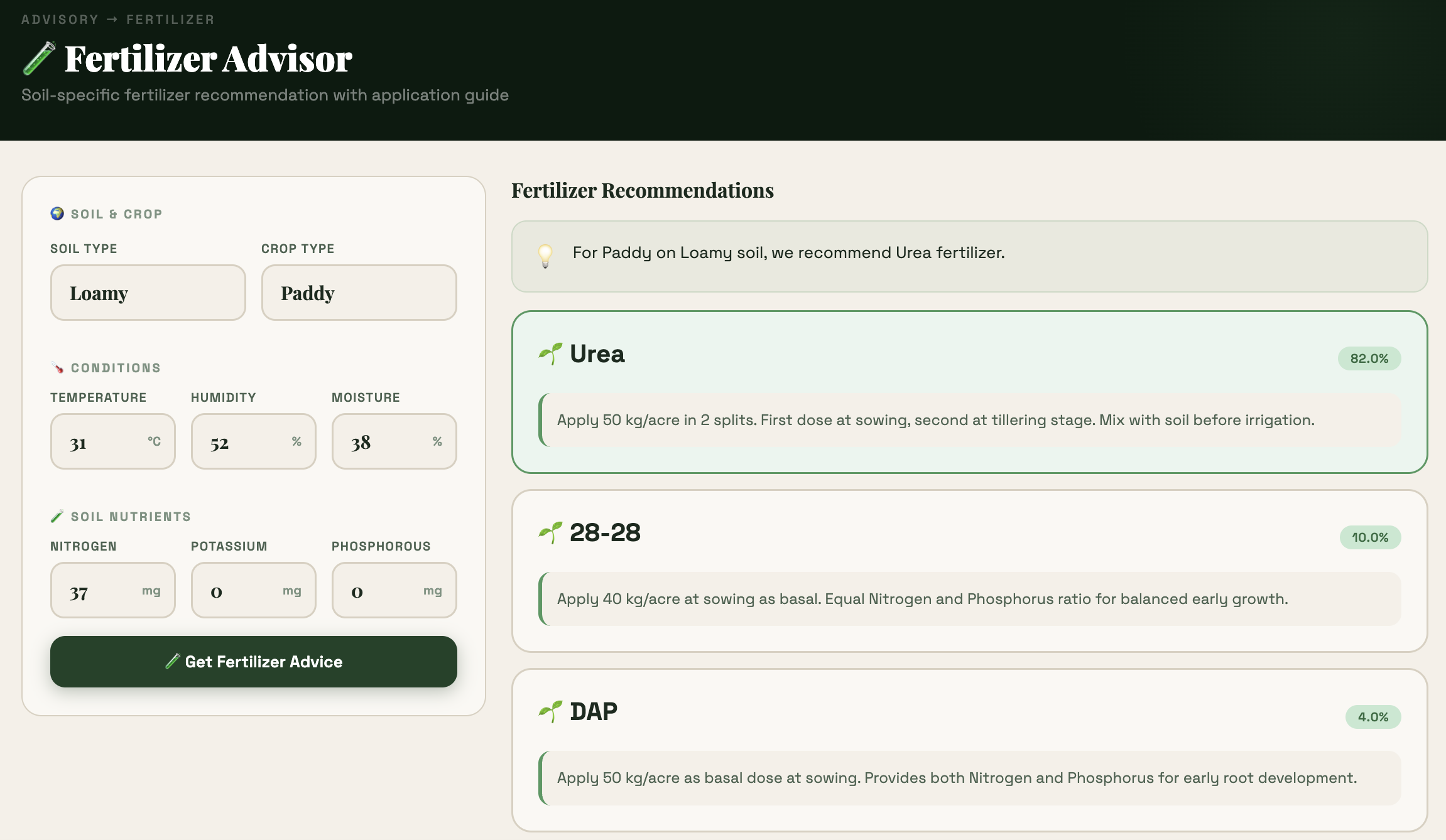}
        \caption{}
    \end{subfigure} 
    \caption{Kisan AI support features providing (a) Crop Advisor and (b) Fertilizer Advisor.}
    \label{fig:flow2}
\end{figure}

\textit{Agronomic Advisory Results}: The Crop and Fertilizer modules implements the RF model. Figure \ref{fig:flow2} shows the capacity of the model to take in soil composition data, including N-P-K content and soil pH level, and give specific crop recommendations and further fertilization advice based on it.

\begin{figure}[h]
    \centering
    \begin{subfigure}{0.47\textwidth}
        \centering
        \includegraphics[width=\linewidth]{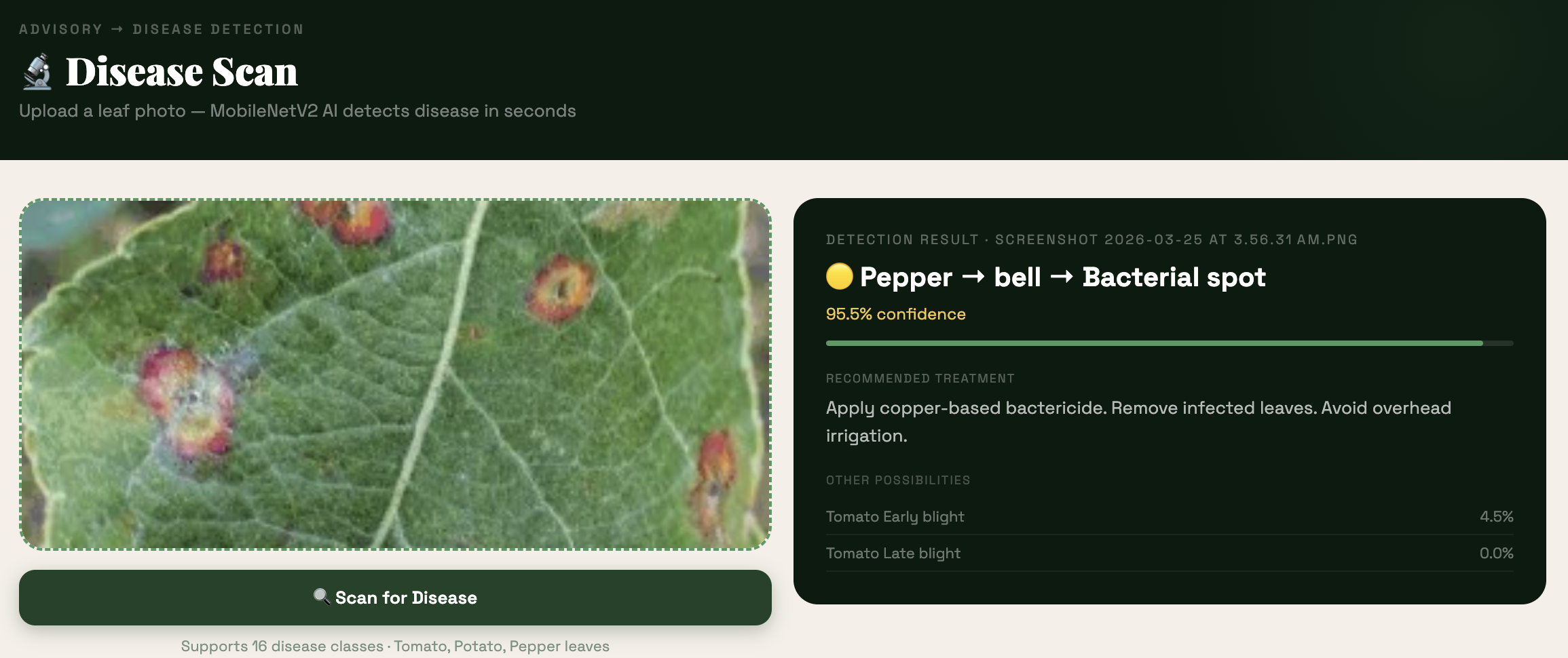}
        \caption{}
    \end{subfigure}
    \begin{subfigure}{0.47\textwidth}
        \centering
        \includegraphics[width=\linewidth]{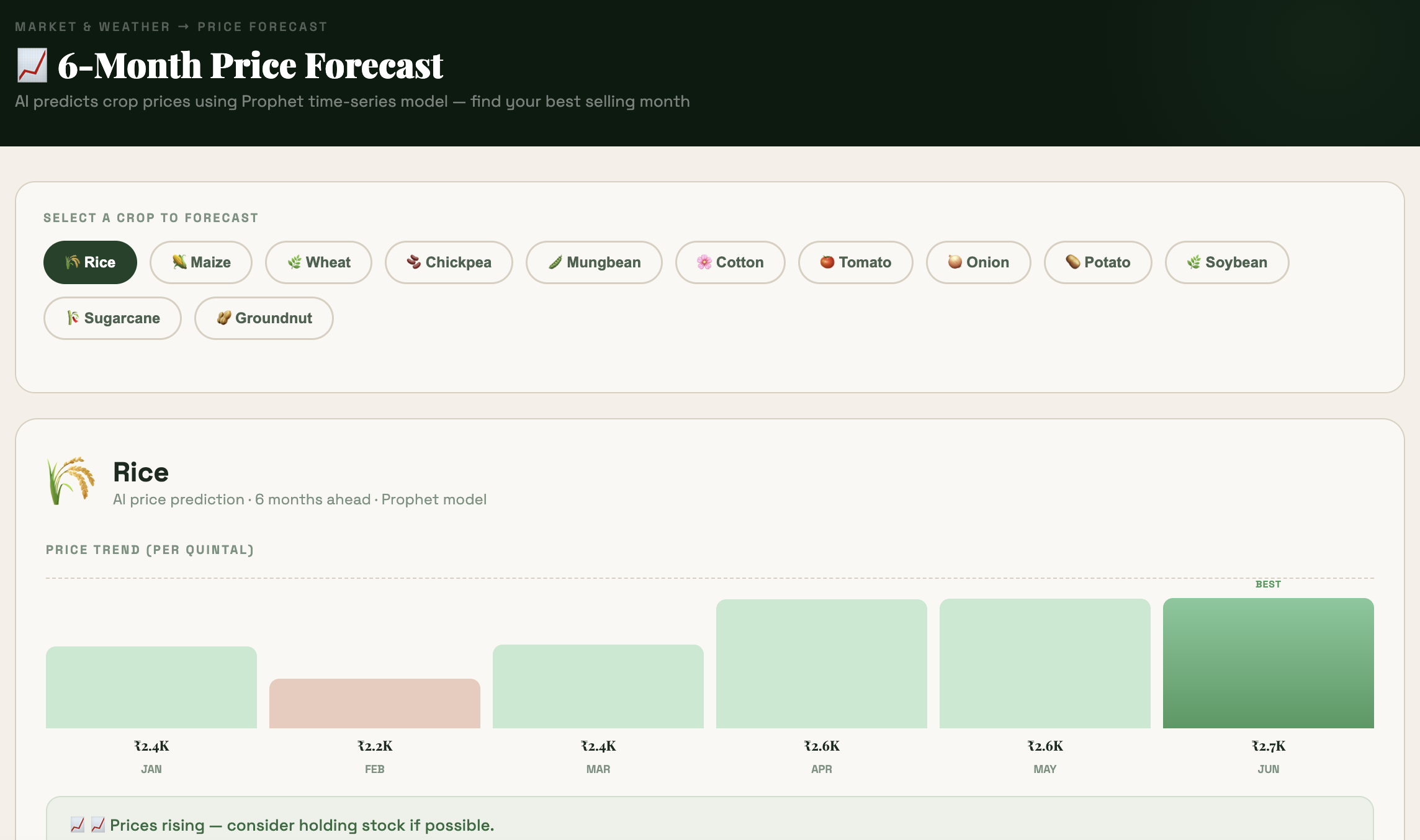}
        \caption{}
    \end{subfigure} 
    \caption{Kisan AI support features providing (a) Disease Detection and (b) Price Trend Analysis.}
    \label{fig:flow3}
\end{figure}

\textit{Pathological Diagnosis and Economic Forecasting}: The vision module, which employs MobileNetV2, reached validation accuracy of 96.2\% on 16 disease categories. Meanwhile, the FB Prophet model generates graphs of the future price trends. In Figure \ref{fig:flow3}, the output of these components is shown; it allows farmers to identify crop leaf diseases through uploading a picture and understand the future price fluctuations of 11 main crop types.

\begin{figure}[h]
    \centering
    \begin{subfigure}{0.45\textwidth}
        \centering
        \includegraphics[width=\linewidth]{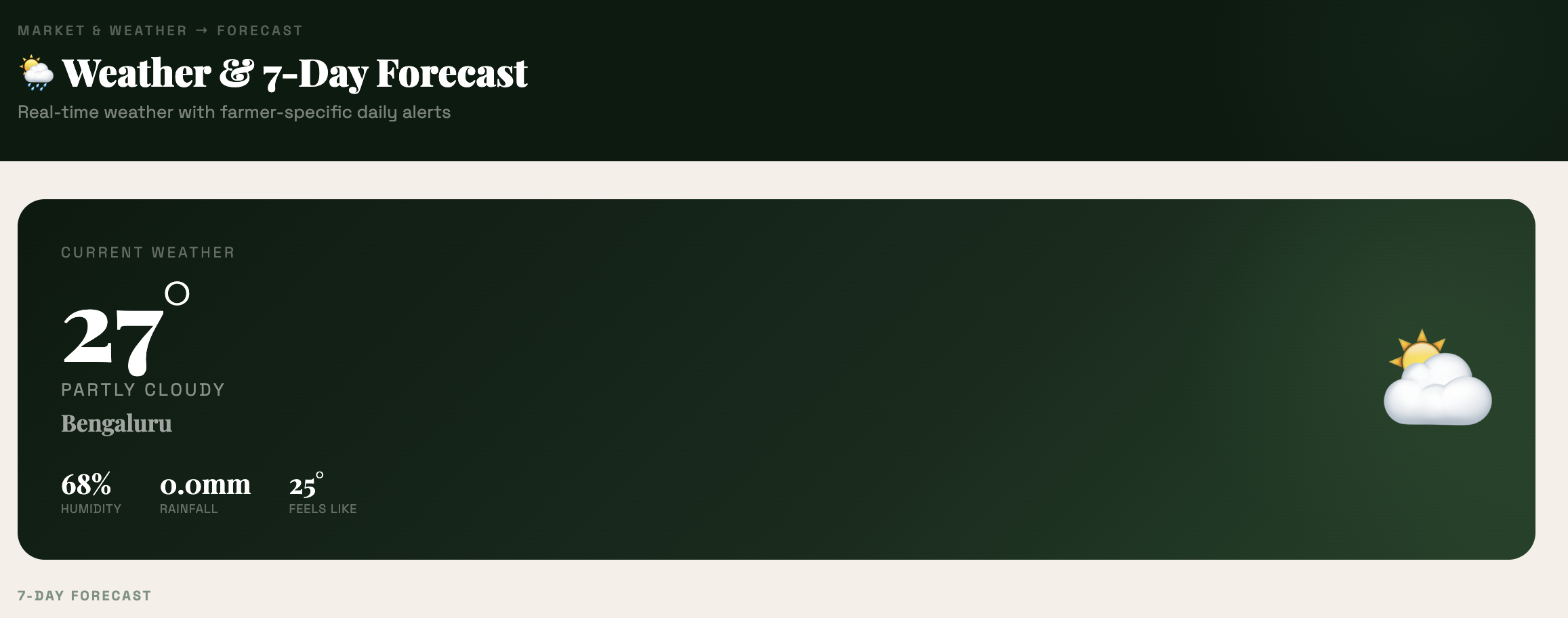}
        \caption{}
    \end{subfigure}
    \begin{subfigure}{0.45\textwidth}
        \centering
        \includegraphics[width=\linewidth]{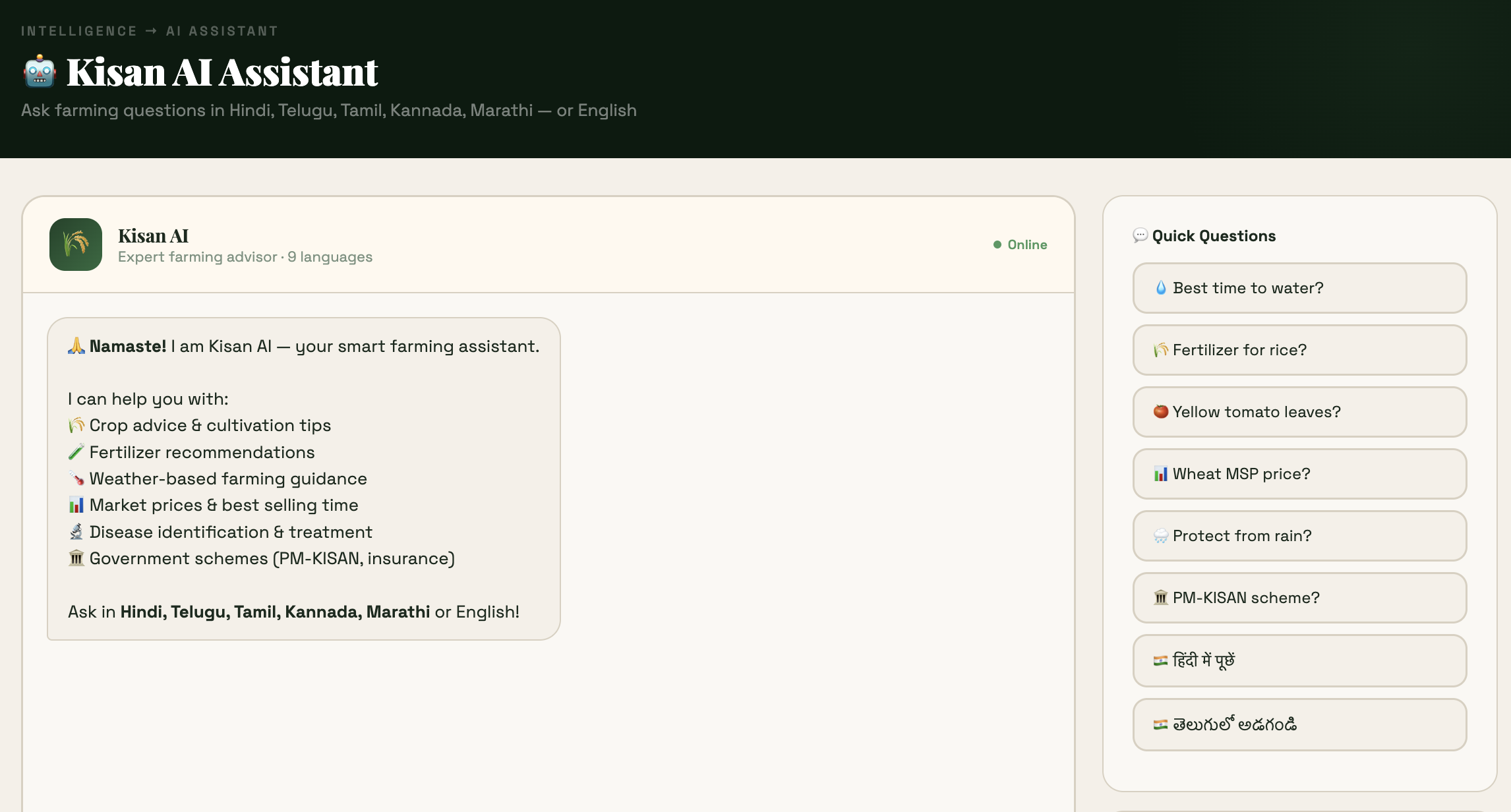}
        \caption{}
    \end{subfigure} 
    \caption{Kisan AI support features providing (a) Weather Alerts and (b) AI Chat Assistant support.}
    \label{fig:flow4}
\end{figure}

\textit{Ancillary Support Features}: To improve the robustness of the advisory system, \textit{Kisan AI} integrates weather alerts and an AI-powered multilingual chatbot to support farmer consultations. As illustrated in Figure \ref{fig:flow4}, these components complement the advisory service by delivering timely, context-aware, and scientifically grounded recommendations across multiple languages.
\subsection{Comparison with the Best-known Baselines}
As demonstrated in Table~\ref{tab:comparison}, the proposed system Kisan AI is either comparable or better than any other previously published results (99.3\% on augmented data). Nevertheless, accuracy is not enough to overcome real-life agricultural problems. The majority of studies have two crucial weaknesses in common; first, the lack of inclusion of economic factors like market price, and second, the inability of deploying the systems. In contrast, the proposed Kisan AI system bridges these gaps by utilizing both approaches at once.

\begin{table}[H]
\centering
\caption{Comparative Analysis of Kisan AI against the Baselines}
\label{tab:comparison}
\small
\begin{tabular}{llccc}
\toprule
\textbf{Study / Year} & \textbf{Model(s) Used} & \textbf{Accuracy} & \textbf{Market Price} & \textbf{Deployed} \\ \midrule
Doshi et al. (2018) \cite{Doshi18} & Naive Bayes & 96.0\% & $\times$ & $\times$ \\
Motamedi et al. (2023) \cite{Motamedi23} & RF + SVM & 99.2\% & $\times$ & $\times$ \\
Senapaty et al. (2024) \cite{Senapaty24} & RF + XGBoost & 98.5\% & $\times$ & $\times$ \\
Gupta et al. (2024) \cite{Gupta24} & Voting Ensemble & 98.5\% & $\times$ & $\times$ \\
Kiran et al. (2024) \cite{Kiran24} & SVM + ANN & 97.2\% & $\times$ & $\times$ \\
Shinde et al. (2025) \cite{Shinde25} & RF + XGB & 98.1\% & $\checkmark$ & $\times$ \\
Shastri et al. (2025) \cite{Shastri25} & ANN + SHAP & 97.8\% & $\times$ & $\times$ \\
\textbf{Kisan AI (Ours)} & \textbf{RF + Prophet} & \textbf{99.3\%} & \boldmath{$\checkmark$} & \boldmath{$\checkmark$} \\ \bottomrule
\end{tabular}
\end{table}

\section{Conclusion and Future Scope}
We developed the research-driven full-stack application \textit{Kisan AI}, a multi-modal crop advisory system to help farmers making agricultural decisions. We selected and implemented the RF model, which achieved $99.3\%$ accuracy for crop prediction, based on our experimental comparisons with state-of-the-art recommendation models. By using market price prediction through FB Prophet and pathology detection through MobileNetV2, our Kisan AI application becomes holistic, transcending the bounds of single-utility agronomic systems.
We will extend the current version of our \textit{Kisan AI} application in the following directions in future.
\begin{itemize}
\item \textit{Multilingual Expansion:} In order to overcome the gap in terms of digital literacy of the target audience, we will include the robust multilingual support.
\item \textit{Transition to Open-Source APIs:} In order to maintain long-term financial stability of the system, it is planned to replace costly third-party APIs with free versions provided either by governments or open sources.
\item \textit{IoT Integration:} We will include an IoT component that allows the system to give dynamic predictions without any user inputs.
\end{itemize}


\begin{thebibliography}{8}
\bibitem{Doshi18} 
Doshi Z, Nadkarni S, Agrawal R, Shah N. AgroConsultant: Intelligent crop recommendation system using machine learning algorithms. In: \textit{IEEE ICCUBEA}, 2018.
\bibitem{Gupta24} 
Gupta S, Sharma R, Jain P. Crop Recommendation System Using ML Algorithms. In: \textit{IRJET Conference Proceedings}, 2024.
\bibitem{Jung23} 
Jung M, Song J S, Shin A Y. Construction of deep learning-based disease detection model in plants. \textit{Scientific Reports}, 2023.
\bibitem{Kiran24} 
Kiran P S, Reddy R, Rajak R K. A system for crop recommendation using SVM and ANN majority voting technique. \textit{International Journal of Recent Technology and Engineering}, 8(211):1125-1130, 2024.
\bibitem{Kumar25}
 Kumar R, Singh A, Patel V. AI-based Smart Crop Recommendation System for Sustainable Agriculture in India. \textit{Madras Agricultural Journal}, 112(4):57-65, 2025.
 \bibitem{Motamedi23} 
Motamedi A, et al. Crop recommendation with fine-tuned SVM and Random Forest. In: \textit{CEUR Workshop Proceedings}, vol. 3992, 2023.
\bibitem{Senapaty24} 
Senapaty M K, Ray A, Padhy N. Decision support system for crop recommendation using machine learning classification algorithms. \textit{Agriculture} 14(8):1256, 2024.
\bibitem{Shastri25} 
Shastri S, Kumar S, Mansotra V. Advancing crop recommendation system with supervised machine learning and explainable artificial intelligence. \textit{Scientific Reports}, 15(25498), 2025.
\bibitem{Shinde25} 
Shinde A D, Patel M J, et al. Ensemble machine learning for crop recommendation and yield prediction. In: \textit{ICCES Conference}, 2025.
\end{thebibliography}
\end{document}